\UseRawInputEncoding 

\documentclass[aps,pra,showpacs,twocolumn, longbibliography]{revtex4-1}
\usepackage{graphicx}
\usepackage{dcolumn}
\usepackage{bm}
\usepackage{color}
\usepackage{amsmath}
\usepackage{amssymb}
\usepackage{setspace}

\begin{document}
\title{Identification of diffracted vortex beams at different propagation distances using deep learning}
\author{Heng Lv$^{1,a}$}
\author{Yan Guo$^{1,a}$}
\author{Zi-Xiang Yang$^{1}$}
\author{Chunling Ding$^{1}$}
\author{Wu-Hao Cai$^{1}$}
\author{Chenglong You$^{2}$}
\email{cyou2@lsu.edu}
\author{Rui-Bo Jin$^{1,3}$}
\email{jrbqyj@gmail.com}
\affiliation{$^{1}$Hubei Key Laboratory of Optical Information and Pattern Recognition, Wuhan Institute of Technology, Wuhan 430205, China}
\affiliation{$^{2}$Quantum Photonics Laboratory, Department of Physics and Astronomy, Louisiana State University, Baton Rouge, Louisiana, USA}
\affiliation{$^{3}$Guangdong Provincial Key Laboratory of Quantum Science and Engineering, Southern University of Science and Technology, Shenzhen, China}
\affiliation{$^{a}$These authors contributed equally to this work.}

\date{\today }

\begin{abstract}
Orbital angular momentum of light is regarded as a valuable resource in quantum technology, especially in quantum communication and quantum sensing and ranging. However, the OAM state of light is susceptible to undesirable experimental conditions such as propagation distance and phase distortions, which hinders the potential for the realistic implementation of relevant technologies. In this article, we exploit an enhanced deep learning neural network to identify different OAM modes of light at multiple propagation distances with phase distortions. Specifically, our trained deep learning neural network can efficiently identify the vortex beam's topological charge and propagation distance with 97\% accuracy. Our technique has important implications for OAM based communication and sensing protocols.
\end{abstract}



\maketitle

\section{Introduction}
Vortex beam generally refers to the phase vortex beam, which has a spiral wavefront, a phase singularity in the center of the beam, and ring-shaped intensity distribution \cite{FrankeArnold2008, RubinszteinDunlop2016}. The beam with orbital angular momentum (OAM) has the phase term $e^{i\ell\phi}$ in the complex amplitude equation, where $\phi$ is the azimuthal angle and $\ell$ is the angular quantum number or topological charge. OAM is an inherent characteristic of vortex beam photons, and each photon carries OAM, which is $\ell\hbar$ \cite{Fickler2012, Bai2022}. Due to the high-dimensional characteristics of the photon OAM, it is utilized in applications such as optical tweezers \cite{Friese1998, Padgett2011}, micromanipulation \cite{Grier2003, Curtis2003}, angular velocity sensing \cite{Lavery2013}, quantum information \cite{Mair2001, MolinaTerriza2007, MaganaLoaiza2019,Ding2015}, quantum computing \cite{Langford2004, Zhang2007, Nagali2009, Zhang2012}, optical communications \cite{Bozinovic2013, Wang2012, Ouyang2021, Zhang2021, Wang2018} and quantum cryptography \cite{MolinaTerriza2001}. Once the value of $\ell$ is identified, the orbital angular momentum can be calculated, allowing the features of the vortex beam to be determined. Unfortunately, the vortex beam will diffract during propagation, and its spatial profile will be easily distorted in a real-world environment \cite{Rodenburg2012}. Detrimentally, the information encoded in the structured beam can be destroyed by random phase distortions \cite{Rodenburg2014, Paterson2005, Tyler2009} and diffraction effects, resulting in mode loss and mode cross-talk \cite{Ndagano2017, Krenn2014}. As a result, capturing the vortex beam and identifying its information using equipment such as a Charge Coupled Device (CCD) camera or a Complementary Metal Oxide Semiconductor (CMOS) camera is difficult \cite{Cox2019}. Hitherto, the traditional methods of identifying vortex beams have included methods such as the interferometer method, plane wave interferometry, and triangular aperture diffraction measurement, to name a few \cite{Courtial1998, Zhang2014,Padgett1996,Yongxin2011}. These traditional methods are much more challenging due to the need for more equipment, as well as complicated data analysis process. In addition, some of these methods can only identify specific vortex beams \cite{Karimi2009}. Moreover, the accuracy of these methods will be greatly reduced when turbulence is considered. These aforementioned factors have significantly hampered the performance of communication, cryptography, and remote sensing. As a result, identifying OAM efficiently and correctly while accounting for diffraction and turbulence is a critical and unresolved challenge.

In recent years, methods such as deep learning algorithm \cite{Shin2016} and transfer learning \cite{Quattoni2008} have considerably increased the accuracy of automatic image recognition \cite{Taigman2014, Melnikov2018}. A significant number of recent articles have proved the potential of artificial neural networks for efficient pattern recognition and spatial mode identification \cite{Doster2017, You2020, Bhusal2021}, and its accuracy is far superior to some traditional identification detection methods \cite{Gibson2004,  Liao2021, Bhusal2021a}.  However, due to the complex diffraction effect in the OAM propagation process, there is little relevant work in the identification of the propagation distance value. In the related research, the propagation distance of the vortex beam ranges from the order of centimeters to the order of kilometers, and it is used as a known parameter \cite{Zhang2019, Krenn2016}. Different propagation distance $z$ will drastically change the size of the central aperture of the vortex beam. As a result, it remains difficult to identify the $z$ value using only the intensity pattern. Additionally, changing the value of topological charge $\ell$ also changes the size of the central aperture, making the identification task more challenging. Finally, turbulence in real-world applications exacerbates the difficulty of such an identification task \cite{Zhao2011}.

In this report, we take advantage of the deep learning algorithm to identify vortex beams and their propagation distances while considering the effects of undesired turbulence. Through theoretical simulations and experiments, we generated vortex beams with different propagation distances and topological charges. In addition, using the transfer learning method, we designed a deep learning model to classify vortex beams. For the first time, our approach utilizes artificial intelligence to simultaneously identify the propagation distance and topological charge of a vortex beam under turbulence's effects. Our research enables the encoding of vortex beams with different propagation distances. As a result, the vortex beam propagation distance may become a new encoding variable. With the improvement of the accuracy of distance recognition, it is even possible to realize precise distance measurement based on the intensity of vortex beams. Our research opens up a new direction for OAM communication and has great significance in OAM based sensing.

\begin{figure*}[!htp]
\begin{center}
\includegraphics[width=0.95\textwidth]{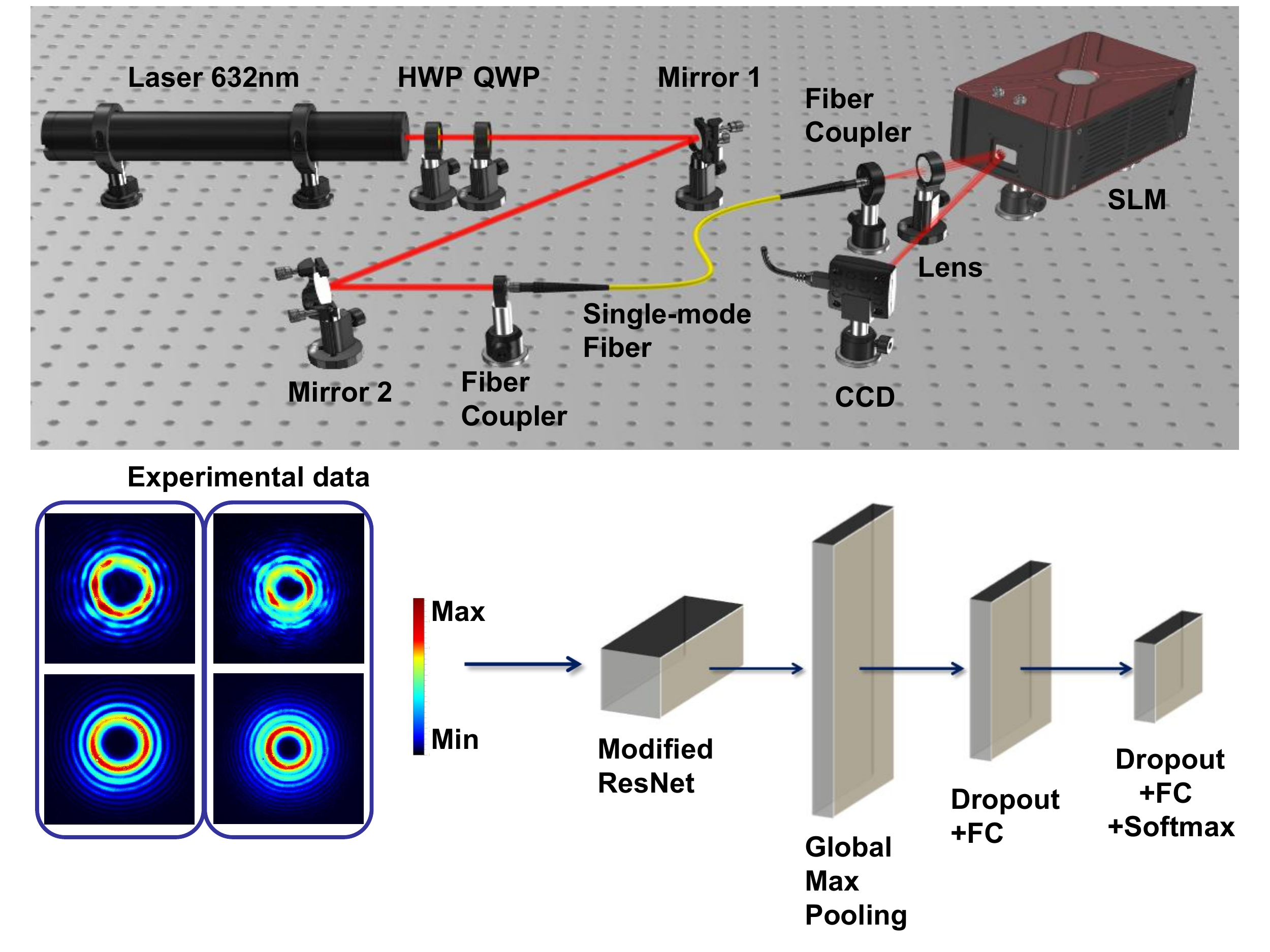}
\end{center}
\caption{The experimental setup (upper panel) and our customized deep learning algorithm (lower panel). We perform the experiment using a collimated He-Ne laser beam. The vortex beam is generated using an SLM with the computer-generated hologram. Finally, the intensity images are collected by a CCD and used for training and testing. Our deep learning network consists of the unaltered ResNet-101 bottom layer and our redesigned top layer. }\label{fig:1}
\end{figure*}

\begin{figure*}[!tbp]
\centering\includegraphics[width=0.95\textwidth]{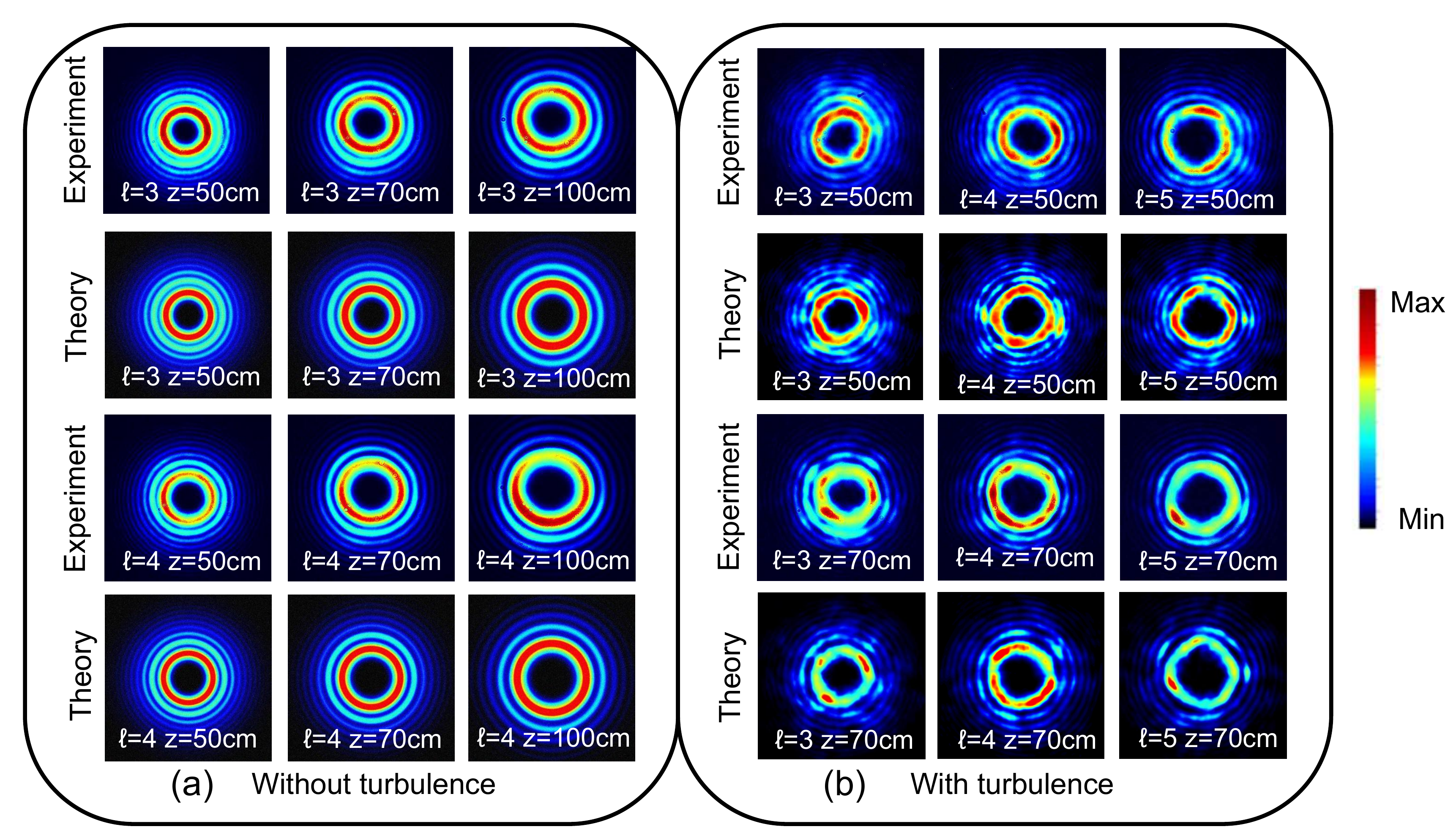}
\caption{The spatial profiles of different vortex beams. We show the experimentally measured images and simulated images for different topological charges $\ell$ and different propagation distances $z$, without turbulence in \textbf{(a)} and with turbulence in \textbf{(b)}. The first and third rows are the images acquired from the experiment, and the second and fourth rows represent the theoretically simulated images.  }
\label{fig:2}
\end{figure*}

\section{Theory and Methods}
\subsection{Generation of the vortex beam}
The fundamental beam used to produce the vortex beam is a Gaussian beam. By applying a phase mask on the spatial light modulator (SLM), the beam amplitude on the plane of SLM becomes \cite{Zhou2016}:
\begin{equation}  \label{eq:1}
E_{1}(r,\theta) =\sqrt{\frac{2}{\pi}}\exp(-\frac{r^{2}}{\omega^{2}_{0}})\exp(-i\ell\theta),
\end{equation}
where $\ell$ is the topological charge, $\omega_{0}$ is the Gaussian beam waist, $r$ and $\theta$ are radial and azimuthal coordinates, respectively.

Within the framework of paraxial approximation, the field distribution of $E_{1}(r,\theta)$ after propagation can be calculated using the Collins integral equation \cite{Collins1970}:

\begin{equation}\label{eq:2}
\begin{aligned}
&E_{2}\left(r_{1}, \theta_{1}, z\right)=\frac{i}{\lambda B} \exp (-i k z) \int_{0}^{2 \pi} \int_{0}^{\infty} E_{1}(r, \theta) \\
&\times \exp \left[-\frac{i k}{2 B}\left(A r^{2}-2 r r_{1} \cos \left(\theta_{1}-\theta\right)+D r_{1}^{2}\right)\right] r d r d \theta,
\end{aligned}
\end{equation}

where $r_{1}$ and $ \theta_{1}$ are radial and azimuthal coordinates in the
output plane, $z$ is the propagation distance, and $k =2\pi/\lambda$ is the wave number with $\lambda$ being the wavelength. The ABCD transfer matrix of light propagation in free space of distant $z$ is
\begin{equation} \label{eq:3}
\left( {\begin{array}{*{20}c}
   A & B  \\
   C & D  \\
\end{array}} \right) = \left( {\begin{array}{*{20}c}
   1 & {z}  \\
   0 & 1  \\
\end{array}} \right).
\end{equation}
By inserting Eq. (\ref{eq:1}) and Eq. (\ref{eq:3}) into Eq. (\ref{eq:2}), we can obtain the beam amplitude as
%
%

\begin{equation}\label{eq:4}
\begin{aligned}
&{E_2}({r_1},{\theta _1},z) = \frac{{{i^{\ell  + 1}}\pi }}{{\lambda z}}\exp \left( { - ikz} \right)\exp \left( { - \frac{{ikr_1^2}}{{2z}}} \right)\\
{\rm{                  }} &\times \exp \left( { - i\ell {\theta _1}} \right)\frac{{b_1^\ell }}{{\varepsilon _1^{1 + \frac{\ell }{2}}}}{\frac{{\Gamma \left( {\frac{\ell }{2} + 1} \right)}}{{\Gamma \left( {\ell  + 1} \right)}}_1}{F_1}\left( {\frac{{\ell  + 2}}{2},\ell  + 1, - \frac{{{b^2}}}{{{\varepsilon _1}}}} \right).
\end{aligned}
\end{equation}

Equation (\ref{eq:4}) represents the hypergeometric Gaussian mode. ${}_{1}F_1(\alpha,\beta,z)$ is a confluent hypergeometric function, $\Gamma (n)$ is the Gamma function, $b_{1}$ and $\varepsilon_{1}$ are defined as:
\begin{equation} \label{eq:5}
{b}_1  = \frac{{kr_1 }}{{2z}},  \quad \varepsilon _1  = \frac{1}{{\omega _0^2 }} + \frac{{ik}}{{2z}}.
\end{equation}
Based on the above calculations, we can obtain transverse intensity images of vortex beam with different values of $\ell$ after propagating different distances $z$.

In actual communication, turbulence can lead to phase distortion of optical mode spatial distribution. Therefore, in our experiment, we use the Kolmogorov model with Von Karman spectrum of turbulence to simulate the atmospheric turbulence in SLM to achieve a distorted communication mode \cite{Bos2015, Glindemann1993}. The degree of distortion is quantified by the Fried parameter $r_{0}$.
The expression of the turbulence phase mask we added on the SLM is \cite{Bhusal2021}:
\begin{equation} \label{eq:6}
\Phi(x,y)=\mathbb{R}\left\{\mathcal{F}^{-1}\left(\mathbb{M}_{N N} \sqrt{\phi_{N N}(\kappa)}\right)\right\},
\end{equation}
with $\phi_{NN} (\kappa) = 0.023r_0^{ - 5/3} (\kappa^2  + \kappa_0^2 )^{ - 11/6} e^{ - \kappa^2 /\kappa_m^2 }$ and the Fried's parameter $r_0  = (0.423k^2C_n^2z)^{- 3/5}$. The  symbol $\mathbb{R}$ represents the real part of the complex field, and  $\mathcal{F}^{-1}$ indicates the inverse Fourier transform operation. In addition, $\kappa$, $\kappa_0$,  and $\mathbb{M}_{N N}$ represent the spatial frequency, the central spatial frequency, and encoded random matrix, respectively. $C_n^2$ is the standard refractive index, which is a constant representing the turbulence intensity.

After the turbulence term is added to the original phase mask and loaded on the SLM,
beam amplitude on the plane of SLM becomes
\begin{equation}  \label{eq:7}
E_{1}^{\prime}(r,\theta)=E_{1}(r,\theta)\exp(i\Phi(x,y)).
\end{equation}
By substituting Eq.(\ref{eq:7})  into  Eq.(\ref{eq:2}), we can numerically obtain the field distribution of the turbulence distorted $E_{2}^{\prime}(r_1,\theta_1, z)$ after propagation.

The experimental setup and the deep learning model are shown in Fig. \ref{fig:1}. In our experiment, the vortex beams are generated by using an SLM and computer-generated holograms. Utilizing the first-diffraction order of SLM, we can obtain the vortex beam of arbitrary topological charge. A laser beam from a He-Ne laser (wavelength of 632.8 nm) is coupled into a single-mode fiber for spatial mode cleaning. A half-wave plate (HWP) and a quarter-wave plate (QWP) are employed to adjust the polarization of the laser beam at the output port of the fiber. An objective lens (magnification of 10$\times$ and an effective focal length of 17 mm) is used to collimate the light from the fiber, and the beam waist after collimation is around 2 mm. By loading a computer-generated phase hologram onto the SLM, a Gaussian beam is converted into a vortex beam. In order to simulate the turbulence in an atmosphere transmission process, we can add an additional turbulence phase to the hologram. Finally, a CCD camera is used to collect the intensity images of the vortex beam, and the transmission distance is controlled by changing the distance between the CCD and the SLM. The images collected by the CCD are sent to a computer for training. Each training set, validation set, and test set contain 86, 10, and 10 images (360 $\times$ 360 pixels), in which the value of $\ell$ ranges from 1 to 5, and the propagation distance $z$ ranges from 40 cm to 100 cm with a step of 5 cm. Totally, there are $86\times5\times13=5590$ images for the training set and $10\times5\times13=650$ images for the validation set and test set.

\subsection{The deep learning algorithm}

The lower panel of Fig. \ref{fig:1} shows our customized deep learning algorithm model. Our model is a transfer learning network based on the ResNet-101 network design \cite{He2016}. Since our obtained images have a high degree of similarity, the neural network must have enough depth to extract image features. Therefore, we adopt the CNN architecture and retrain the ResNet-101 deep learning model rather than the shallow neural networks model. More specifically, the top layer is removed from the original ResNet-101. Moreover, a global max pooling layer with a node count of 2048 is used to reduce the parameters to increase the calculation speed. Following that, a dropout layer is added to remove some parameters randomly to minimize over-fitting, and then we use a fully connected (FC) layer to connect the local features. Another dropout layer and an FC layer are added to lower the number of nodes from 1024 to 65. Finally, a softmax layer is applied for a 65 classification probability output.

To train and test the deep learning model, we utilize a computer with an Intel(R) Core(TM) i5-7300HQ CPU @2.5GHz and an Nvidia GeForce GTX 1050 Ti GPU with 4GB of video memory. We use an adaptive moment estimation (Adam) optimizer throughout the algorithm \cite{Kingma2014}. In our deep learning model, we also used the transfer learning technique (TLT) \cite{Fernando2014}, which has two benefits. Firstly, it is highly efficient; for example, tasks that originally required months of training without TLT can be reduced to a few hours. The second merit of TLT is that less data is needed. This is because transfer learning requires the use of a pre-trained model, which allows us to achieve accurate recognition results with fewer datasets \cite{Patel2015}. Generally speaking, only hundreds or thousands of training images (instead of tens of thousands or even millions of images) are enough to achieve good training results \cite{Krizhevsky2017, Liu2019}. Finally, the training results in each epoch are evaluated by the categorical cross-entropy loss function \cite{10.5555/3327546.3327555} which is given by
\begin{equation}
\text{ Loss }=-\sum_{i=1}^{n}\left(\hat{y}_{i 1} \ln y_{i 1}+\hat{y}_{i 2} \ln y_{i 2}+\ldots+\hat{y}_{i m} \ln y_{i m}\right),
\end{equation}
where $n$ is the number of samples, $m$ is the number of classifications, $\hat{y}_{i m}$ indicates that the true label (with the value of 0 or 1), and $y_{im} $ is the predicted value of the $m$-th class given by the neural network.

\section{Results and discussion}

\begin{figure*}[!htbp]
\begin{center}
\includegraphics[width=15cm]{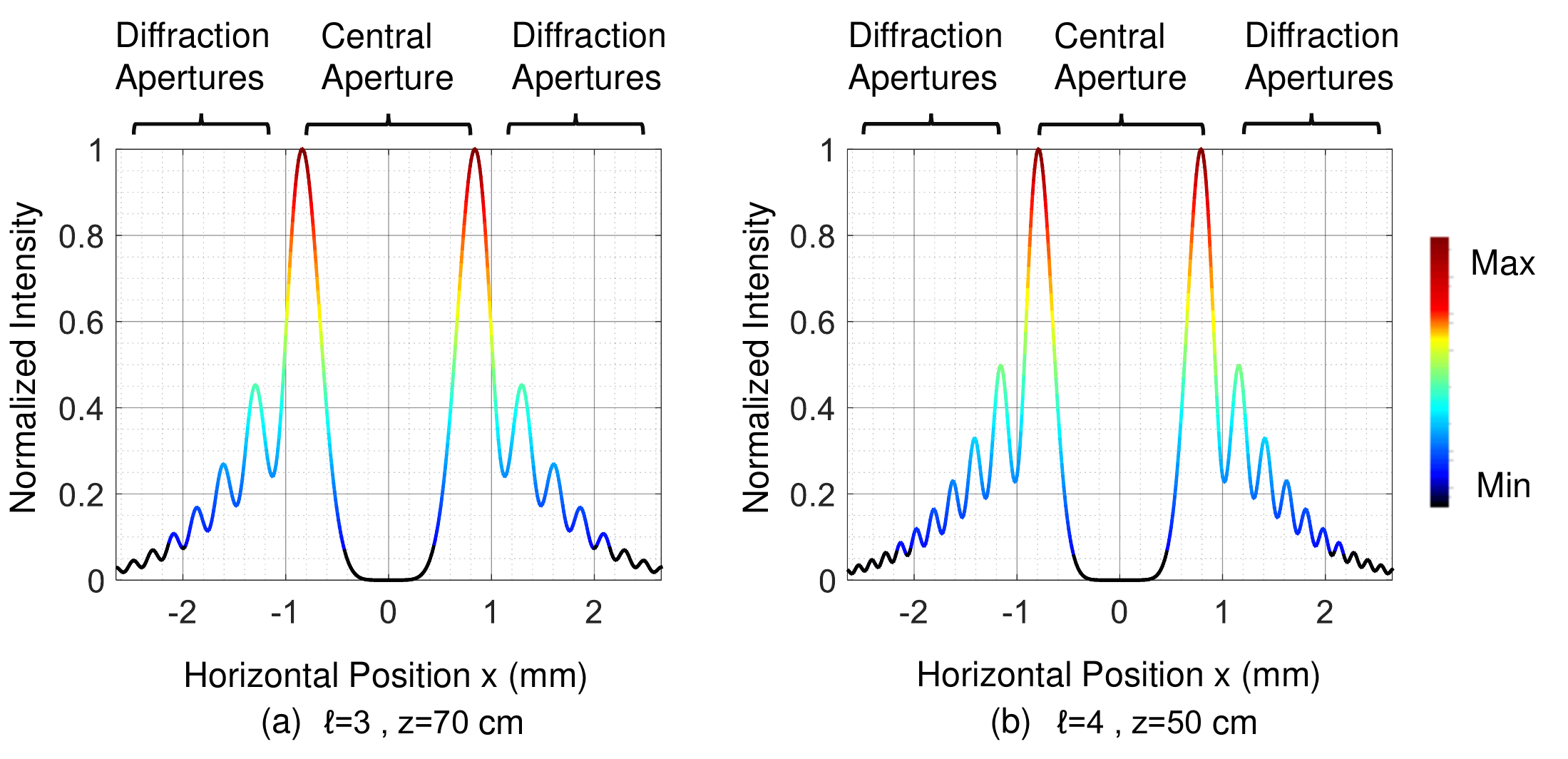}
\end{center}
\caption{The cross-sectional view of the intensity images at $y=0$ for \textbf{(a)} $\ell=3$, $z=70$ cm  and \textbf{(b)} $\ell=4$, $z=50$ cm. The size of the central apertures is comparable in both images. However, the number of side lobes in the diffractive apertures is different. This subtle feature enables us to distinguish these two cases.}
\label{fig:3}
\end{figure*}

\begin{figure*}[!thp]
\begin{center}
\includegraphics[width=1\textwidth]{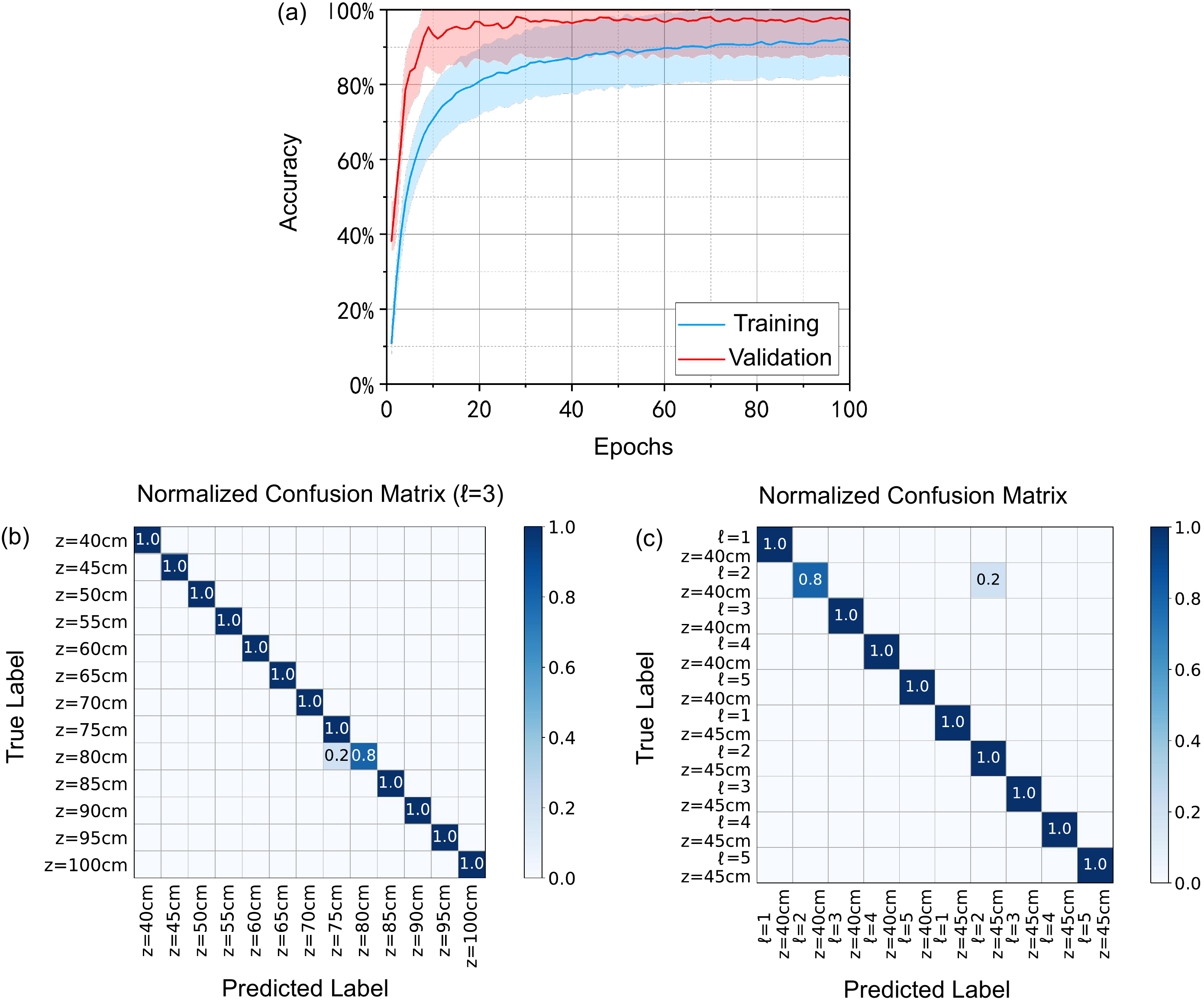}
\caption{The accuracy and confusion matrix of our trained deep learning algorithm. \textbf{(a)} The accuracy of the training set and the validation set versus the epochs. The accuracy of up to 97\% is achieved in identifying vortex beams with different $\ell$ values and different $z$ values after 90 epochs. All shaded areas correspond to the standard deviation of accuracy.  All classification results have been tested and verified, and some test results are shown here. \textbf{(b)} The normalized confusion matrix between the predicted propagation distance and the true propagation distance for $\ell=3$. \textbf{(c)} Normalized confusion matrix between predicted $\ell$ values, predicted propagation distance and true $\ell$ values, true propagation distance.}
\label{fig:4}
\end{center}
\end{figure*}

Figure \ref{fig:2} shows the spatial profiles of vortex beams with different propagation distances $z$ and topological charges $\ell$ obtained from experiments and simulations. Figure \ref{fig:2}(a) shows the vortex beams without turbulence, and Fig. \ref{fig:2}(b) depicts vortex beams affected by turbulence. The first and third rows depict the spatial profiles of the vortex beams acquired in the experiment, while the second and fourth rows depict the simulated ones under identical conditions. For simplicity, we define the first aperture in the center as the ``center aperture" and other outer apertures as the ``diffraction apertures". For a fixed value of $\ell$, the size of the central aperture becomes larger as the propagation distance $z$ value increases. At the same time, for a fixed propagation distance $z$, the size of the central aperture also increases as the value of $\ell$ increases. From Fig.  \ref{fig:2}, we can observe that the spatial profiles obtained from the experiment match well with our simulations, therefore validating our theoretical model of the experiment. By comparing the experimental and theoretical images of $\ell=4$ and $z=100$ cm, we can notice that the central aperture in the experimental image is not distributed uniformly. This effect is due to the slight misalignment of the collimated beam to the center of the SLM. As a result, the brightness and shape of the center aperture can vary slightly. This kind of deviation is also included, in order to increase the diversities of the training data. We will show later that, even with such deviations, the training results remain excellent.

There are many situations where the size of the central aperture is comparable for vortex beams with different $\ell$ and propagation distance $z$. This particular effect makes the simultaneous identification of $\ell$ and $z$ difficult. For example, by comparing the theoretical image of $\ell=3$ and $z=70$ cm (second row, second column), and the image of  $\ell=4$ and $z=50$ cm (fourth row, first column) in Fig. \ref{fig:2}(a), we can notice that the sizes of the center aperture are similar, making it difficult to distinguish between these two modes. In this case, the difference of the diffraction apertures provides the best characteristic value to distinguish them. More intuitively, Fig. \ref{fig:3} shows the cross-sectional view of the intensity for these two beams at $y=0$. It is clear that the distance between the two main peaks in Fig. \ref{fig:3}(a) and Fig. \ref{fig:3}(b) is almost the same. However, Fig. \ref{fig:3}(a) has 4 side lobes in the diffraction apertures, while Fig. \ref{fig:3}(b) has 6 side lobes in the diffraction apertures. This subtle difference makes it possible to distinguish these two cases. Finally, we note that in Fig. \ref{fig:3} the diffraction apertures in the region of $|x|>2.2$ mm (in black) usually cannot be well captured by a CCD in the experiment, due to low light intensity and limited resolution of the CCD.

In practical communication applications, the spatial profile of the vortex beam might be distorted due to atmospheric turbulence, underwater turbulence, or other adverse circumstances. Therefore, the turbulence should be taken into account for the propagation of the vortex beam. Figure \ref{fig:2}(b) shows the typical simulated and experimental diagrams with turbulence. In these theoretical (the second and fourth rows) and experimental (the first and third rows) data, a turbulence intensity parameter of  $C_{n}^{2}=5\times10^{-10}$  mm$^{-2/3}$ is utilized. In these data sets, we also take into account the fact that light might not be precisely incident on the center of the SLM plane. It can be noticed that the turbulence created huge distortions on the vortex beam's center aperture and diffraction apertures. For deep learning training, we gathered 1040 distorted light intensity images.

To show the performance of our deep learning network, we plot the accuracy as a function of the training epochs in Fig. \ref{fig:4}(a). We can see that after 90 training epochs, the accuracy is higher than 92\% for the training set, while the accuracy of the validation set is greater than 97\%. Since we added regularization and dropout operations during the training process. These operations will be automatically closed during the verification process, causing the accuracy of the validation set to be higher than the accuracy of the training set. A high validation accuracy of 97\% indicates that our approach provides a powerful way to identify the vortex beams with different $\ell$ values and different $z$ values, even under a turbulent environment. Finally, we note that the number of epochs required for convergence depends on multiple factors, including the number of cases of different vortex beams propagated and the degree of turbulence.

To show our results more comprehensibly, we calculated the normalized confusion matrix for different $\ell$ and $z$. Figure \ref{fig:4}(b) shows a typical normalized confusion matrix from $\ell=3$, $z=40$ cm to $\ell=3$, $z=100$ cm.  Figure \ref{fig:4}(c) shows the normalized confusion matrix from $\ell=1$, $z=40$ cm to $\ell=5$, $z=45$ cm. The true propagation distance and the predicted propagation distance given by our deep learning algorithm are basically on a diagonal line. The result means almost all OAM modes and propagation distances tested are correctly identified, only two images with $\ell=3$, $z=80$ cm are predicted to be $\ell=3$, $z=75$ cm in Fig. \ref{fig:4}(b), and only two images with $\ell=2$, $z=40$ cm are predicted to be $\ell=2$, $z=45$ cm in Fig. \ref{fig:4}(c).

Finally, we want to emphasize that using classical methods (e.g., interferometer) to analyze the distorted intensity images in Fig. \ref{fig:2}(b) is quite challenging. However, according to training and test results, our deep learning model can accurately identify vortex beams with varying topological charges and propagation distances even under the influence of severe turbulence. This demonstrates that our approach has a high level of robustness and is very useful for practical applications. We note that our approach can be adapted to identify larger $\ell$ value with longer and more accurate transmission distance. However, due to the limitation of our equipment, such as the resolution of SLM and CCD, as well as the experimental error caused by the laboratory environment, we limit the size of our topological charges and the length of the propagation distance. We believe our designed deep learning neural network does not fundamentally limit the recognition accuracy, and its potential is far from being reached. Moreover, our scheme can be adapted to many vortex beam related applications. For instance, we can adapt our work to consider multiple types of vortex beams, and even the combination of them. Furthermore, the accurate identification of the propagation distance might be a novel technique for sensing related applications. Last but not least, our approach can be applied to free-space OAM communication, especially the demodulation system, to increase the robustness of the communication. We expect that by combining the unique characteristics of vortex beams with the advantages of the deep learning algorithm, more breakthroughs in vortex beams research can be made in the future.

\section{Conclusion}
Vortex beams have enormous potential due to their versatility and virtually unlimited quantum information resources. However, these beams are highly susceptible to undesirable experimental conditions such as propagation distance and phase distortions. In our work, we exploit the deep learning algorithm to identify a vortex beam's topological charge and propagation distance. Specifically, we focus on vortex beam with topological charge $\ell$ from 1 to 5, and the propagation distance $z$ ranges from 40 cm to 100 cm. Additionally, we consider the effect of turbulence-induced in the propagation of the beam. We experimentally demonstrated that our customized deep learning algorithm could accurately identify the propagation distance and topological charge. Our work has important implications for the realistic implementation of OAM-based optical communications and sensing protocols in a turbulent environment.

\section*{Conflict of Interest Statement}
The authors declare that the research was conducted in the absence of any commercial or financial relationships that could be construed as a potential conflict of interest.

\section*{Author Contributions}
The idea was conceived by R.-B.J.,  C.Y., and H.L. The experiment was designed by H.L., Y.G., C.Y., and R.-B.J. The experiment was performed by H.L. with help of Y.G., Z.X.Y, C.D., W.-H.C.,  C.Y. and R.-B.J.  The theoretical description and numerical simulation were developed by Y.G., C.Y., and R.-B.J. The  deep learning process was carried out by H.L.  The data was analyzed by H.L., Y.G., C.Y., and R.-B. J.  The project was supervised by R.-B.J. and C.Y. All authors contributed to the preparation of the manuscript.

\section*{Funding}
This work is supported by the National Natural Science Foundations of China (Grant Nos.12074299, 91836102,11704290) and by  the Guangdong Provincial Key Laboratory (Grant No. GKLQSE202102).

\section*{Acknowledgments}
We thank Prof. Zhi-Yuan Zhou for the helpful discussion.

\end{document}